\documentclass{appolb}
\usepackage{epsf}
\usepackage[dvips]{color}

\begin{document}

\pagestyle{plain}

\title{Linguistic complexity: English vs. Polish, text vs. corpus}

\author{Jaros{\l}aw~Kwapie\'n$^1$, Stanis{\l}aw~Dro\.zd\.z$^{1,2}$, Adam Orczyk$^1$ \address{$^1$ Institute of Nuclear Physics, Polish Academy of Sciences, Krak\'ow, Poland, \\
$^2$ Faculty of Mathematics and Natural Sciences, University of Rzesz\'ow, Rzesz\'ow, Poland}}

\maketitle

\begin{abstract}

We analyze the rank-frequency distributions of words in selected English and Polish texts. We show that for the lemmatized (basic) word forms the scale-invariant regime breaks after about two decades, while it might be consistent for the whole range of ranks for the inflected word forms. We also find that for a corpus consisting of texts written by different authors the basic scale-invariant regime is broken more strongly than in the case of comparable corpus consisting of texts written by the same author. Similarly, for a corpus consisting of texts translated into Polish from other languages the scale-invariant regime is broken more strongly than for a comparable corpus of native Polish texts. Moreover, we find that if the words are tagged with their proper part of speech, only verbs show  rank-frequency distribution that is almost scale-invariant.


\end{abstract}

\section{Introduction}

Even though central to the contemporary science the concept of complexity - by its very nature - still leaves its precise definition an open issue. In intuitive and qualitative terms this concept refers to diversity of forms, to emergence of coherent and orderly patterns out of randomness, but also to a significant flexibility that allows switching among such patterns on a way towards searching for the ones that are optimal in relation to environment. Physics offers the concepts that seem promising for formalizing complexity, among which is criticality implying a lack of characteristic scale which indeed finds evidence in abundance of power laws and fractals in Nature.

Whatever definition of complexity one however adopts, the human language deserves a special status in the related investigations. It not only led humans to develop civilization but it also constitutes - from a scientific perspective - an extremely fascinating and complex dynamical structure~\cite{nowak00}. Like many natural systems, language during its evolution developed remarkable complex patterns of behaviour such as hierarchical structure, syntactic organization, long-range autocorrelations, and - what is particularly relevant here - a lack of characteristic scale. This latter phenomenon - in quantitative linguistics commonly referred to as the Zipf law - describes the rank-frequency distribution of words in a (sufficiently large) piece of text. This well-known, quantitatively formulated in 1949 by G.K. Zipf observation~\cite{zipf49}, based originally on ``Ulysses'' and latter on confirmed for many other literary texts, states that frequency of the rank-ordered words is inversely proportional to the words' rank. It needs to be added here that this law constitutes a principal reference in quantitative linguistics and inspiration for ideas and development in many different areas of science.

Zipf suggested interpretation of this law in terms of the so-called principle of least effort, stating that words are used in such a manner that their frequency is optimized with respect to the cost of using them~\cite{zipf49,mandelbrot54,ferrer03}. This interpretation was however soon questioned after it had been shown that the Zipfian relation applies also to a ``typewriting monkey'' example~\cite{miller57}, i.e. an essentially purely random process. This pointed to the Zipf law as too indiscriminate to reflect the complex organization of languages. Our recent studies based on English as well as on Polish texts shed some new light on an involved message encoded in the Zipf law.

We analyze the rank-frequency distributions of words based on selected literary texts and small corpora. We concentrate on differences between texts written in English and in Polish, on differences between texts written by the same or by different authors, and on differences between native and translated texts. We also tag words with their corresponding parts of speech (classes) and consider the rank-frequency distributions of words representing only a specific class.

\section{Methods}

The Zipf's Law in its original form can be expressed by
\begin{equation}
f(r) \sim r^{-\alpha}
\label{zipf}
\end{equation}
where $r$ denotes a word's rank and $f(r)$ is the number this word occurs in a sample. The scaling index $\alpha$ for English is on average above 1~\cite{montemurro01}, but can slightly vary for individual authors and works. It can also assume different values for other languages. Despite its simplicity, the relation~(\ref{zipf}) describes the empirical data quite well except for the most frequent words, which actually tend to be underrepresented. (These most frequent words can also be comprised in an analytical distribution if one of the generalizations of the Zipf's Law, e.g. the Zipf-Mandelbrot's Law is applied, but this is beyond the scope of the present paper.)

The convention for forming a ranking is that the word which occurs most frequently has rank 1, the second most frequent word has rank 2, and so on. Typically, before the ranking can be calculated, each piece of text has to be preprocessed in order to erase all the non-word strings (like numbers or special-character strings). Then we eliminate notable spelling errors by comparing each string with entries of a dictionary. Both tasks are done automatically by a computer. However, in order to assure that no valid words are in this way modified (e.g. neologisms and deliberately misspelled words), we also check the erased words manually. Finally, all the words are counted within a sample (text or corpus). If a scale-invariant behaviour of $f(r)$ is present, we estimate the index $\alpha$ via regression.

Any text, as a piece of human syntactic communication, is not a series of grammatically equivalent words, but rather a mixture of interwoven words belonging to different classes. Tagging all the words with their proper class allows us to compare statistical properties of words within each class separately. In this case, words belonging to the inflectable classes - nouns, verbs, adjectives and pronouns - can be considered twofold. They can be counted either in their actual inflected forms as they appear in a text or in their lemmatized (basic) forms, e.g. the infinitive in the case of verbs and the nominative singular in the case of nouns. The latter approach is worth attention due to the fact that in such a way we can obtain a better (but not one-to-one) mapping between words and concepts, and minimize the influence of grammar.

We perform lemmatization in parallel with part-of-speech tagging. This is the most demanding part of our analysis, especially for English words which typically are flexible and can act as different parts of speech (in Polish such situations are less common and are easier to handle). We do not apply any tagging software, which at present does not offer 100\% reliability, but instead we accomplished this task semi-automatically (i.e. automatic tagging of unambiguous words and manual tagging of the rest of words).

\section{Results}

\begin{figure}
\hspace{-0.5cm}
\epsfxsize 7cm
\epsffile{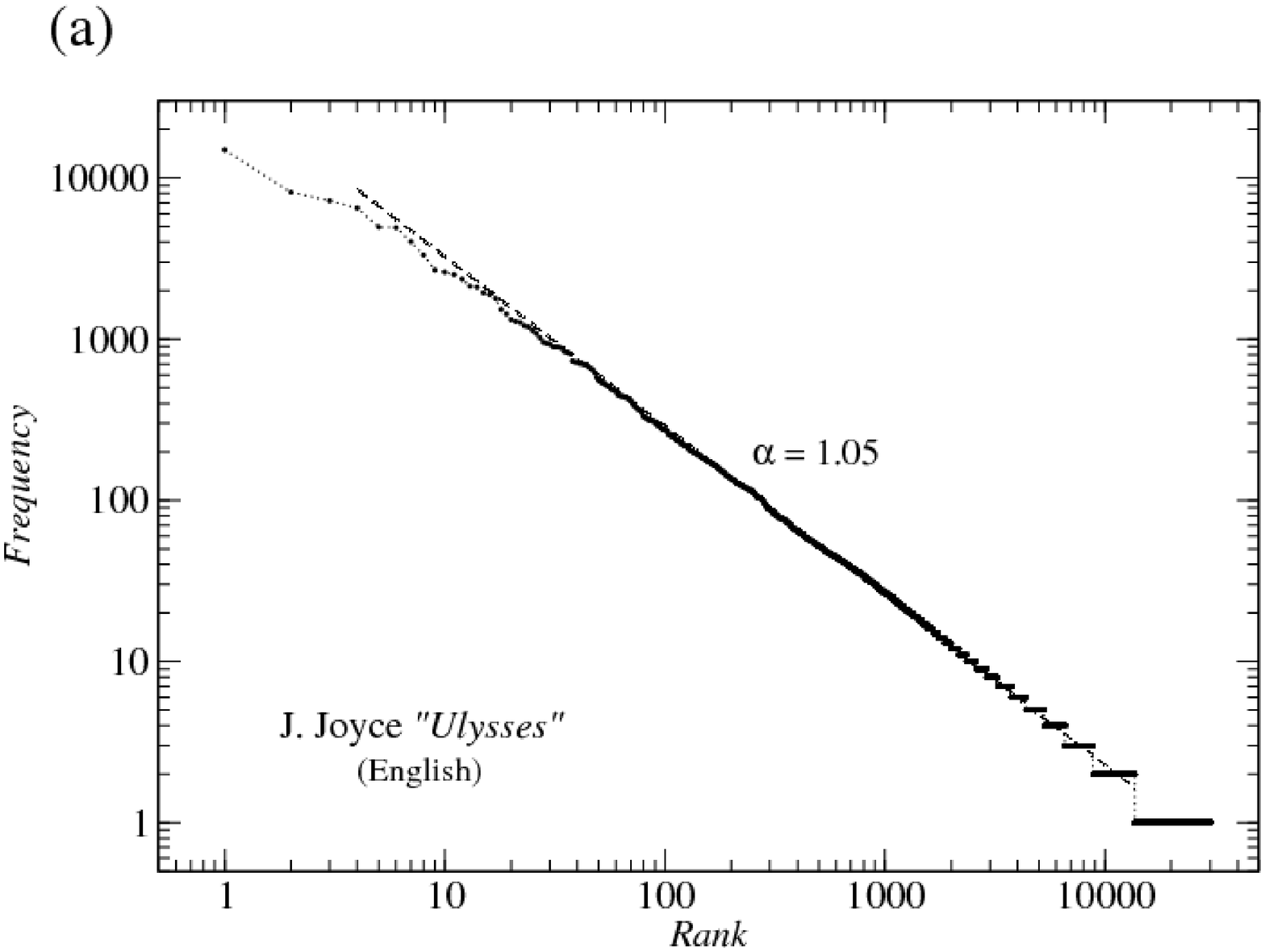}
\epsfxsize 7cm
\epsffile{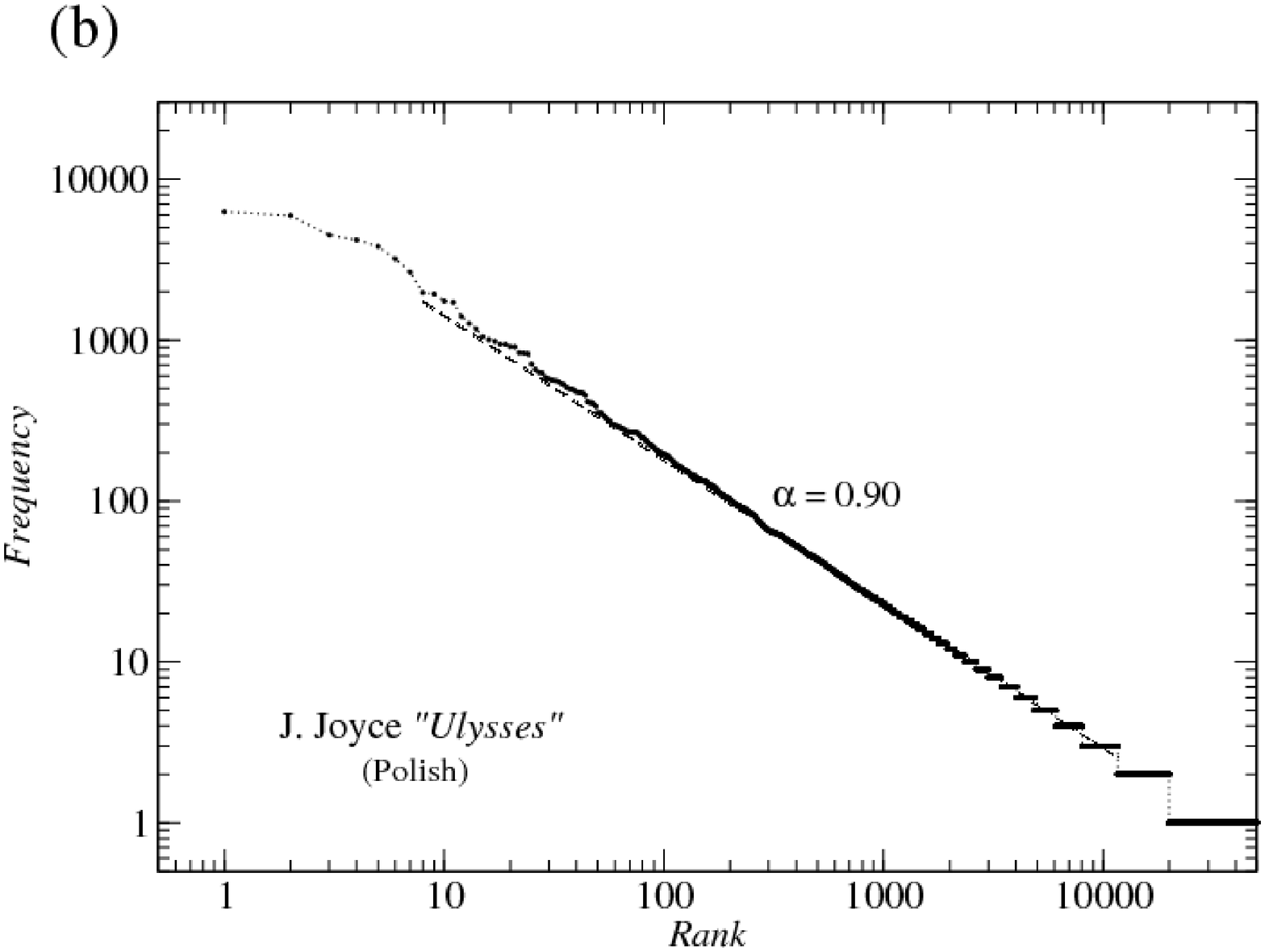}

\hspace{-0.5cm}
\epsfxsize 7cm
\epsffile{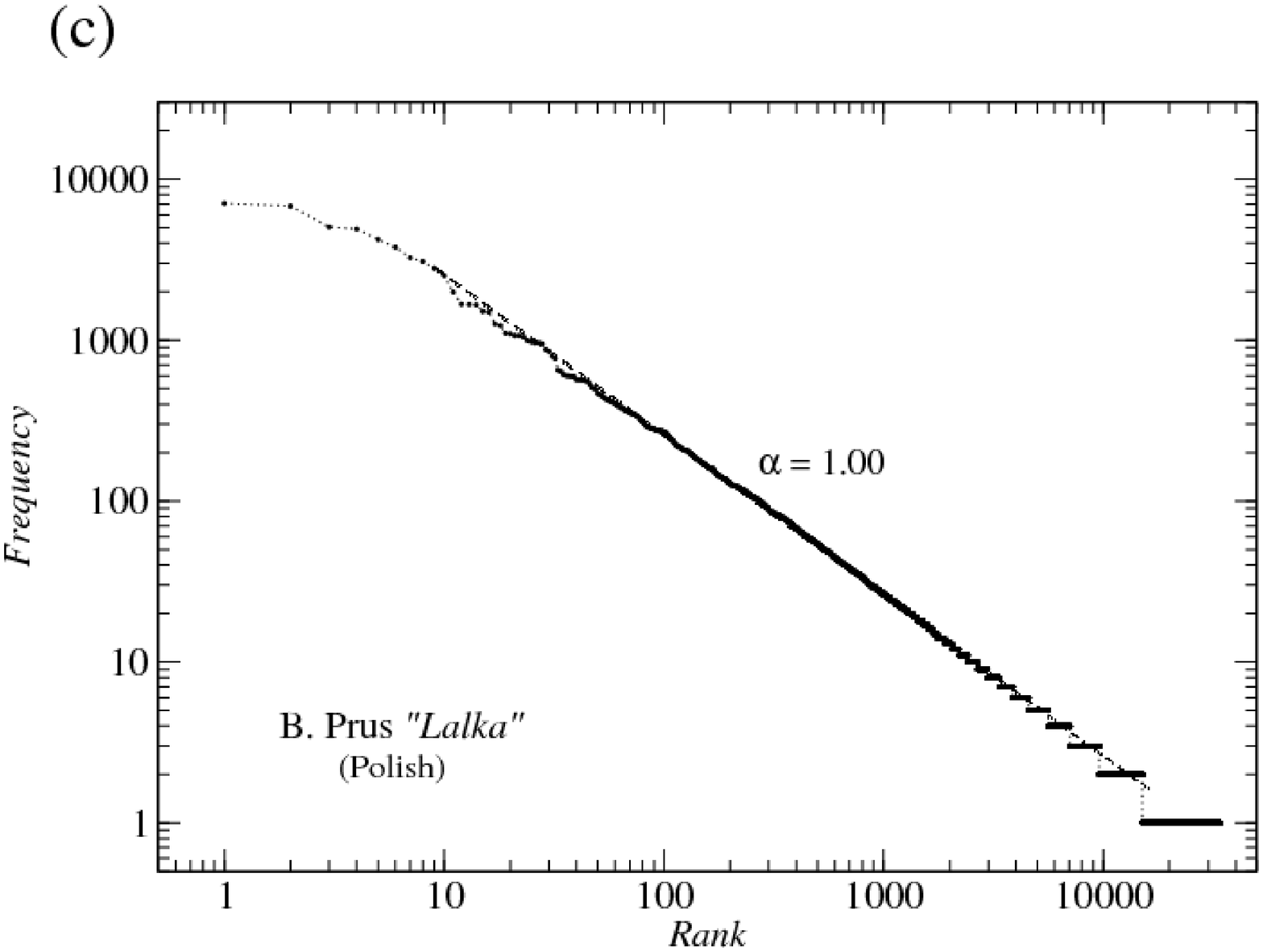}
\epsfxsize 7cm
\epsffile{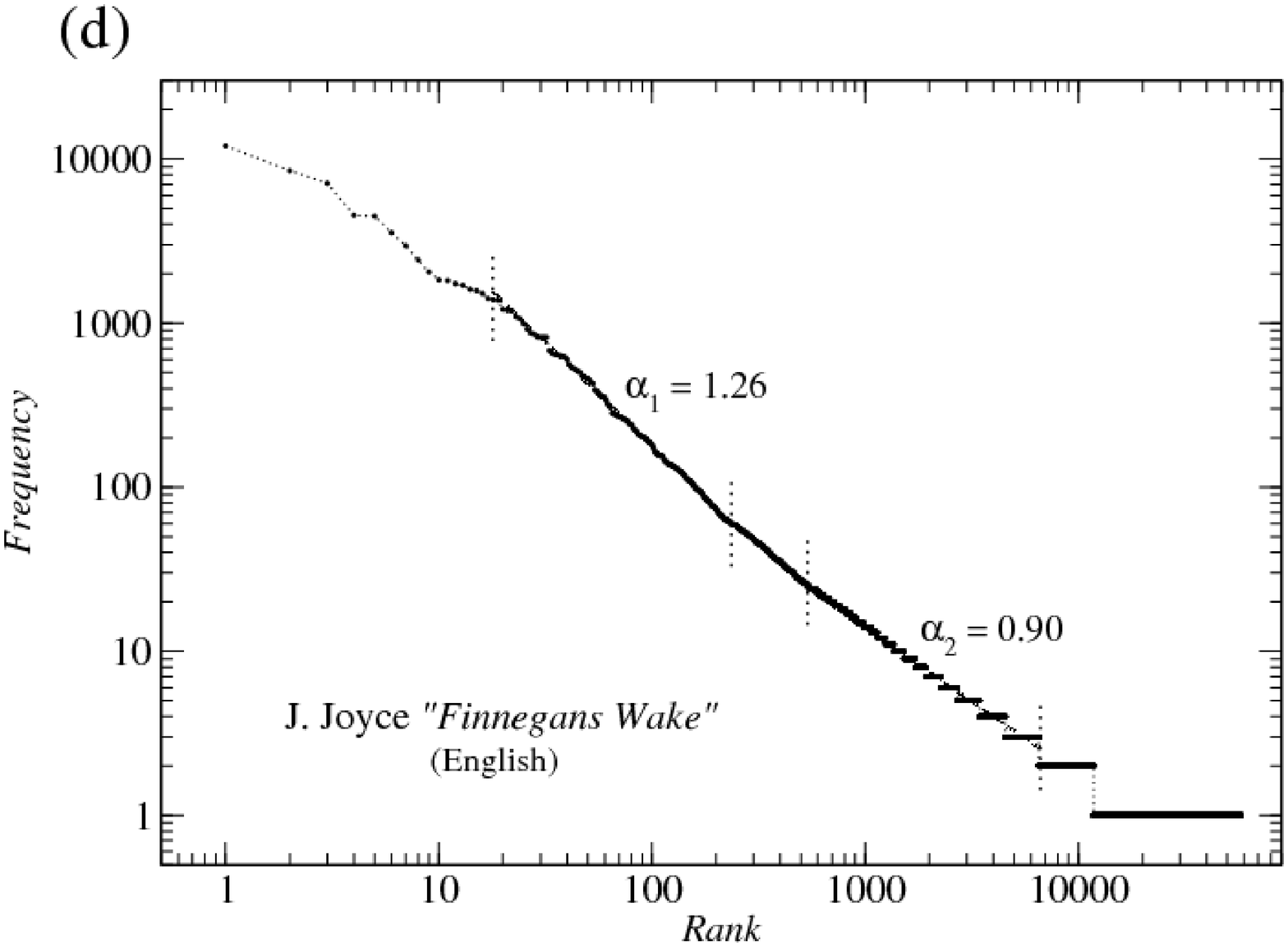}
\caption{Rank-frequency distributions of words in the English (a) and Polish (b) text of ``Ulysses'' by J.~Joyce, as well as in the native Polish text of ``Lalka'' (``The Doll'') by B.~Prus (c) and the English text of ``Finnegans Wake'' by J.~Joyce (d). Note the lack of a unique scale-invariant regime and the overrepresentation of rare words in (d) as compared to (a)-(c).}
\end{figure}

First, referring to the classic analysis of ``Ulysses'' by James Joyce, carried out by G.K.~Zipf~\cite{zipf49}, we compare properties of the rank-frequency distributions of words in the English and the Polish text~\cite{slomczynski} of this novel. English and Polish represent different groups of the Indo-European languages and their grammars are considerably different, which makes such a parallel study instructive. Results are shown in Figure 1(a) and Figure 1(b). Both versions of the text show scale-invariant behaviour over three decades between the ranks 10 and 10,000. However, the power-law exponent for the Polish text ($\alpha \simeq 0.90$) is smaller than for the English original ($\alpha \simeq 1.05$). This difference might originate from a far more inflectable character of the Polish language, which demands a larger set of words (understood as particular sequences of letters) to reproduce the course of narration in ``Ulysses''. This observation may be considered a regularity, since typical Polish texts have smaller $\alpha$ than typical English texts (on average: $~0.95$ vs. $~1.05$~\cite{montemurro01}). There are, however, Polish texts which have a value of $\alpha$ that is similar to English standards. An example of such a text is the 19th century Polish novel ``Lalka'' (``The Doll'') by Boles{\l}aw Prus, as it is documented in Figure 1(c). We selected this particular novel for the present analysis due to its length (245,292) comparable to that of the English text of ``Ulysses'' (264,272). It is noteworthy that each of the three texts are well approximated by power-law distributions almost over the whole range of ranks - a property that is rarely realized in typical texts, which at higher ranks usually show a more or less pronounced downward deflection from the power law due to a limited vocabulary volume of a writer. From this point of view, a consistent power-law behaviour of high-ranked words one may consider to be a merit of a writer who equally easily operates common and specific vocabulary. However, there are exceptions from this general regularity. Figure 1(d) shows a Zipf plot created for the J.~Joyce's novel ``Finnegans Wake''. Here no overall scale-invariant behaviour can be observed, but instead the two short (one order of magnitude) power-laws can be identified with $\alpha_1=1.26$ for the ranks 20-200 and $\alpha_2=0.90$ for the ranks 500-6000. A curious property of this novel, which is opposite to other texts, is the flattening of the curve with increasing rank; this property extends even over the words that occur only once. ``Finnegans Wake'' is famous for being a highly experimental piece of text and this stylistic artificiality is predominantly responsible for its unusual statistical properties. Despite its uncommon properties, this novel seems to be an interesting subject for future studies due to the fact that Joyce here tried to follow a flow of thoughts and visions one experiences during sleep ("stream of consciousness"). Thus, its language might be less artificial and unnatural as it might seem at first glance.

\begin{figure}
\hspace{-0.5cm}
\epsfxsize 7cm
\epsffile{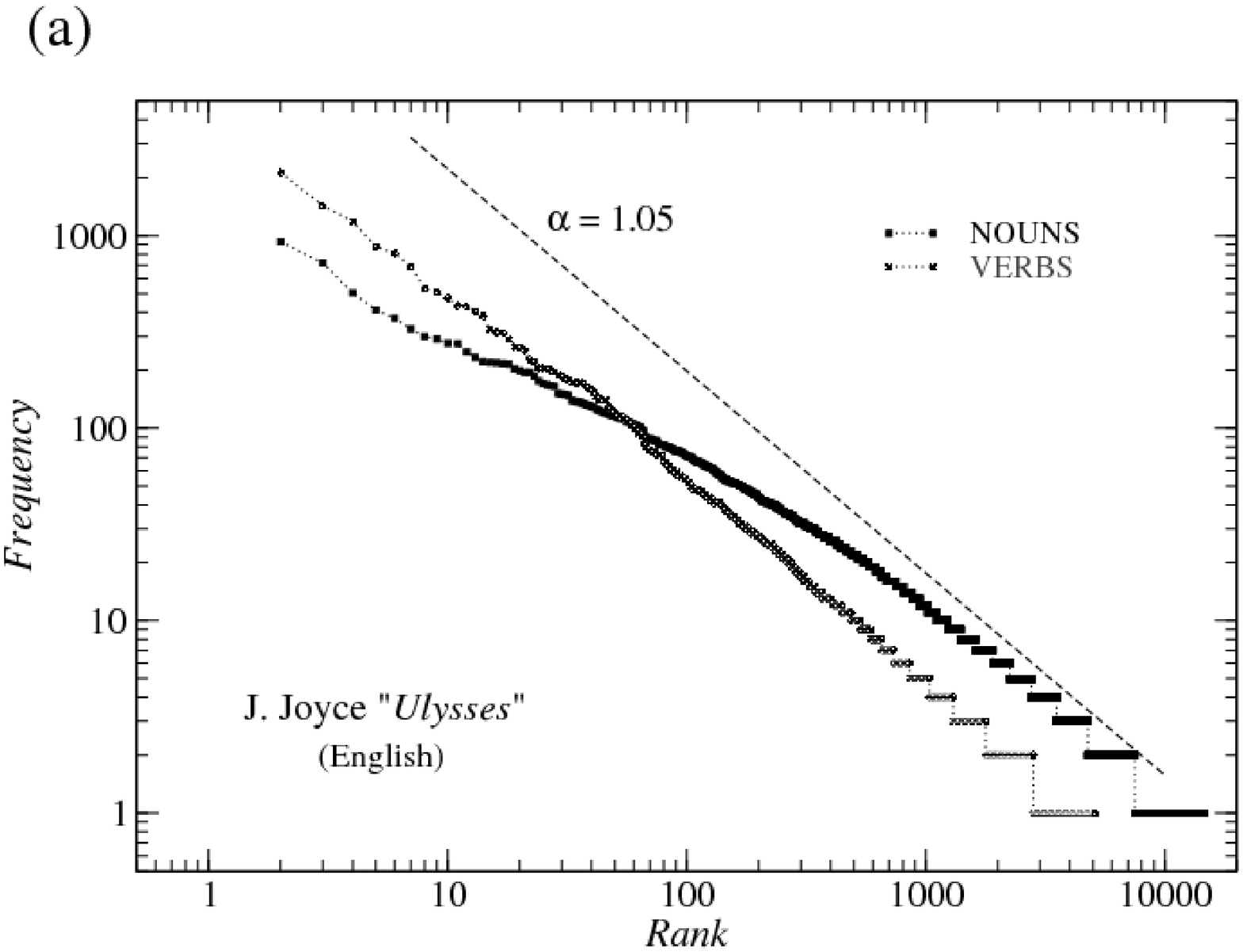}
\epsfxsize 7cm
\epsffile{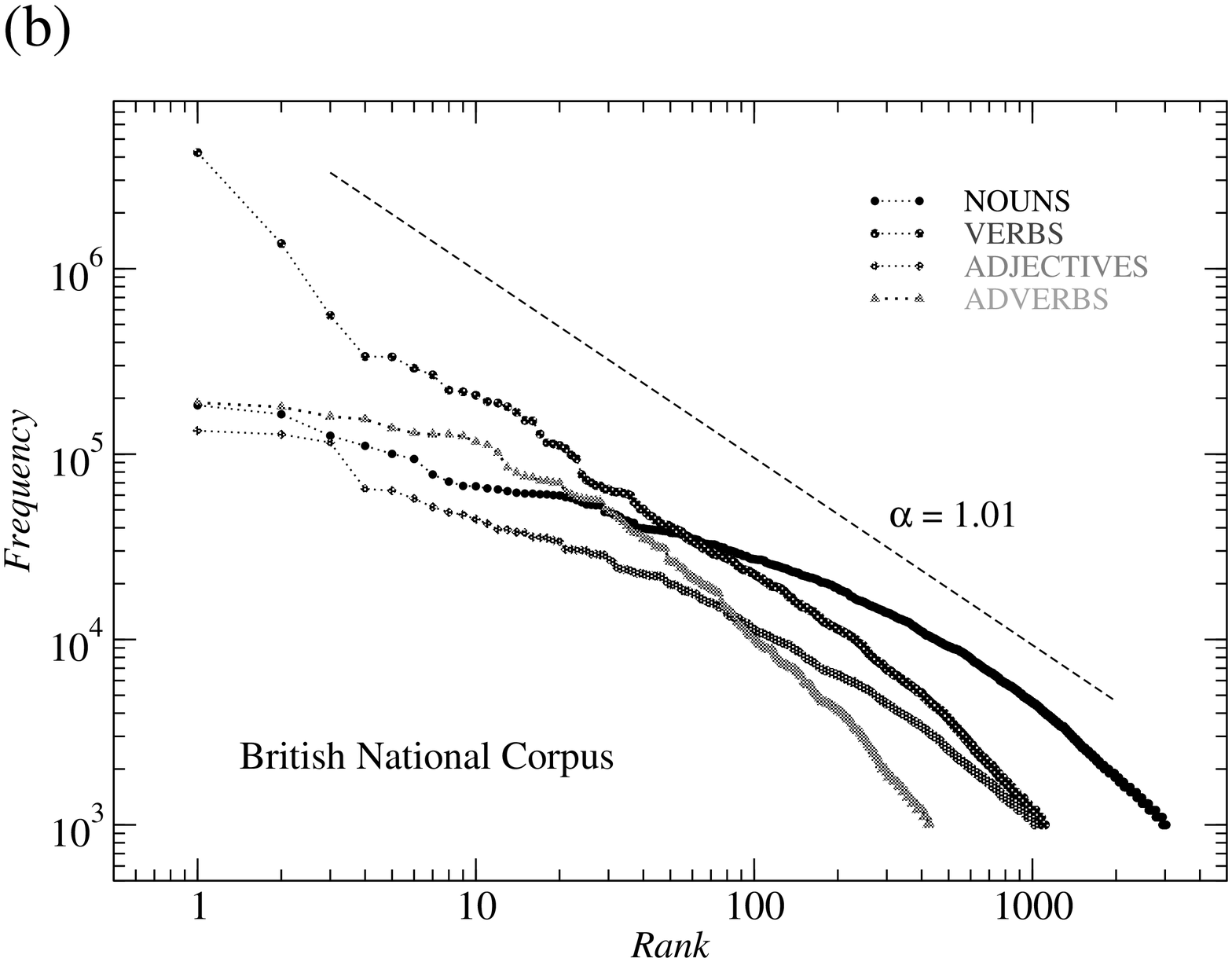}

\hspace{2.5cm}
\epsfxsize 7cm
\epsffile{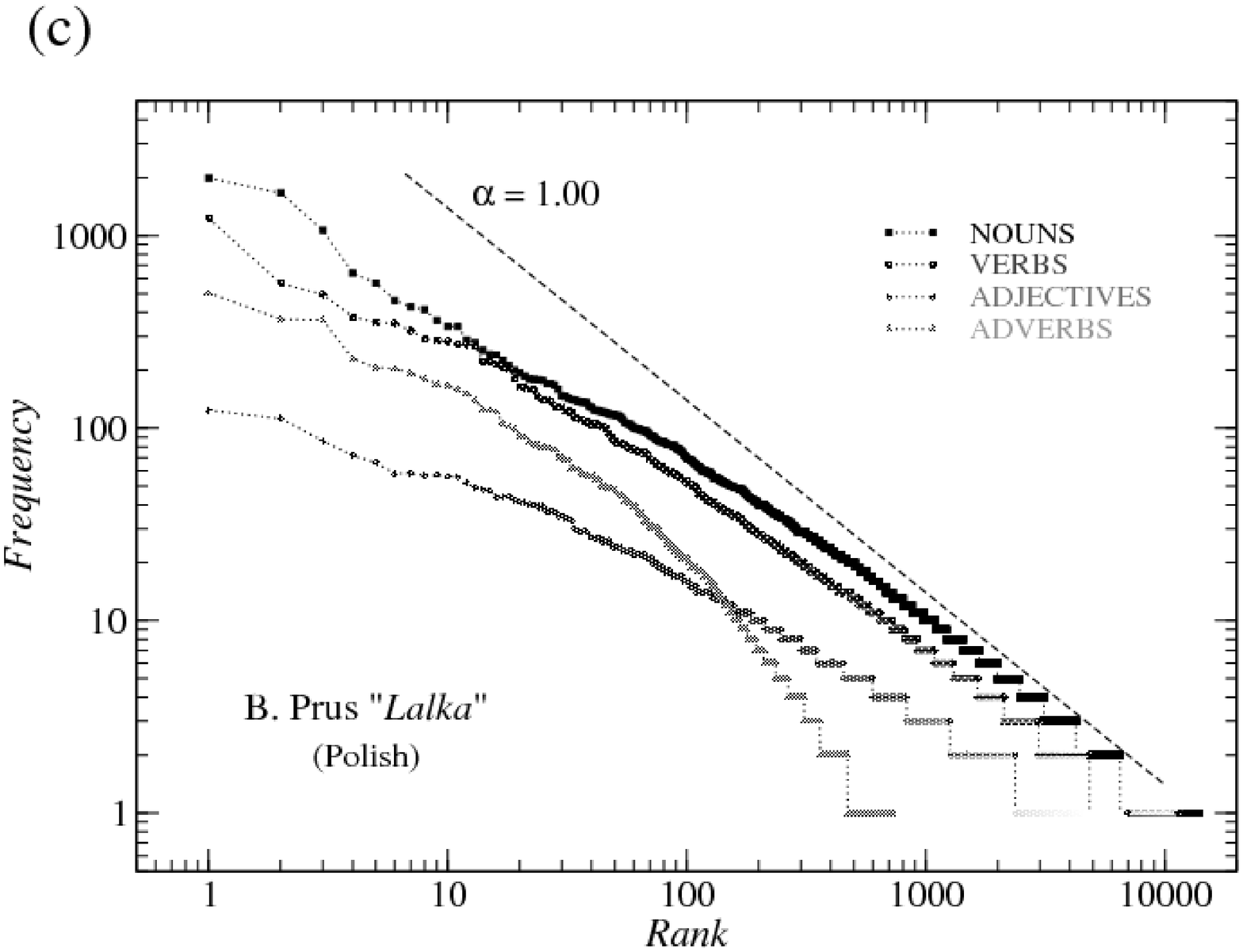}
\caption{Class-specific rank-frequency distributions of words in the English text of ``Ulysses'' (a), in the British National Corpus (b), and in the Polish text of ``Lalka'' (c). Word classes are nouns (black squares), verbs (dark grey circles), adjectives (medium grey diamonds), and adverbs (light grey triangles). In each panel a slope with $\alpha$ equal to its value for the corresponding complete vocabulary is shown as a reference. Note: in the case of ``Ulysses'', the part-of-speech tagging was restricted to nouns and verbs.}
\end{figure}

Figure 2(a) shows the rank-frequency distributions of the most frequent nouns and verbs in ``Ulysses''. Despite the fact that the complete texts exhibit power-law dependence (Figure 1(a)), the corresponding distributions may not necessarily show such behaviour if only words representing a single class are considered. Unlike nouns, verbs are a class that presents some trace of scale-invariant character with $\alpha \approx 1$. In order to simplify the task, while tagging the text of ``Ulysses'', we concentrated primarily on nouns and verbs and instead of distinguishing between adjectives, adverbs and the remaining 4 word classes, we considered such words to be a black-box class. Therefore we cannot present the corresponding results in Figure 2(a). However, the approximate sub-rankings, formed by extracting out of the global ranking the words which typically function as adjectives and adverbs, indicate that words from neither of these two classes show scale invariance.

That these are not specific properties of ``Ulysses'', but they are more generally realized, one can infer from Figure 2(b), where the corresponding results are shown for the British National Corpus~\cite{bnc}, a representative sample ($10^8$ words) of the contemporary written and spoken English (the data come from~\cite{leech}). In this corpus, which as a whole shows a Zipf-type scale invariance with $\alpha=1.01$ for ranks up to several thousands~\cite{ferrer01}, no class exhibits a power-law, but these are verbs which are the closest to it. We have not carried such an analysis for a representative Polish corpus, so we can present only the results for ``Lalka'' (Figure 2(c)). As in English, the Zipf plots for the selected word classes differ in this case from the complete vocabulary shown in Figure 1(c) and do not present scale invariance. However, a difference between verbs and nouns is not so prominent as in Figures 2(a) and 2(b) yet still verbs are the closest to a power-law.

\begin{figure}
\hspace{-0.5cm}
\epsfxsize 7cm
\epsffile{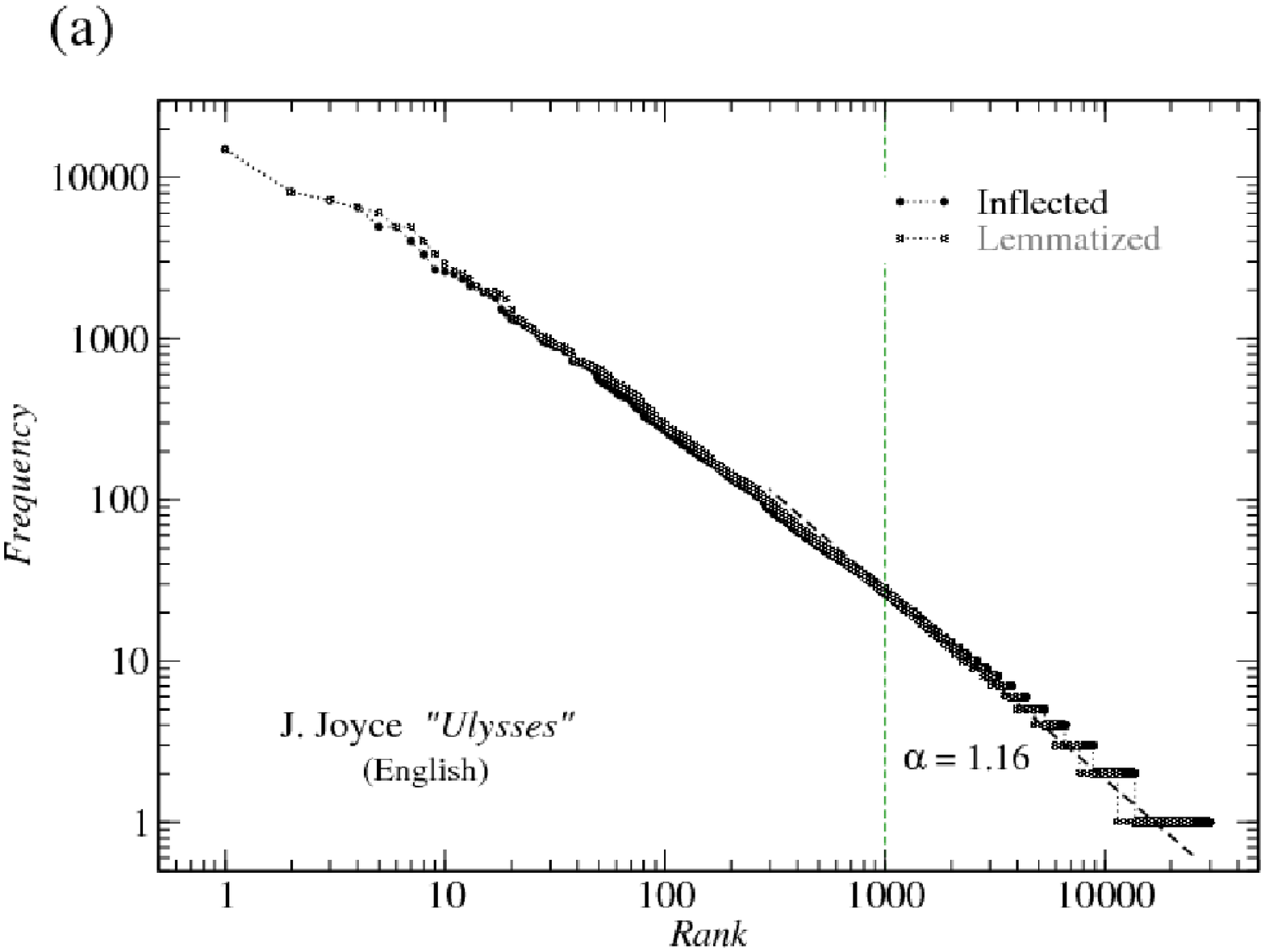}
\epsfxsize 7cm
\epsffile{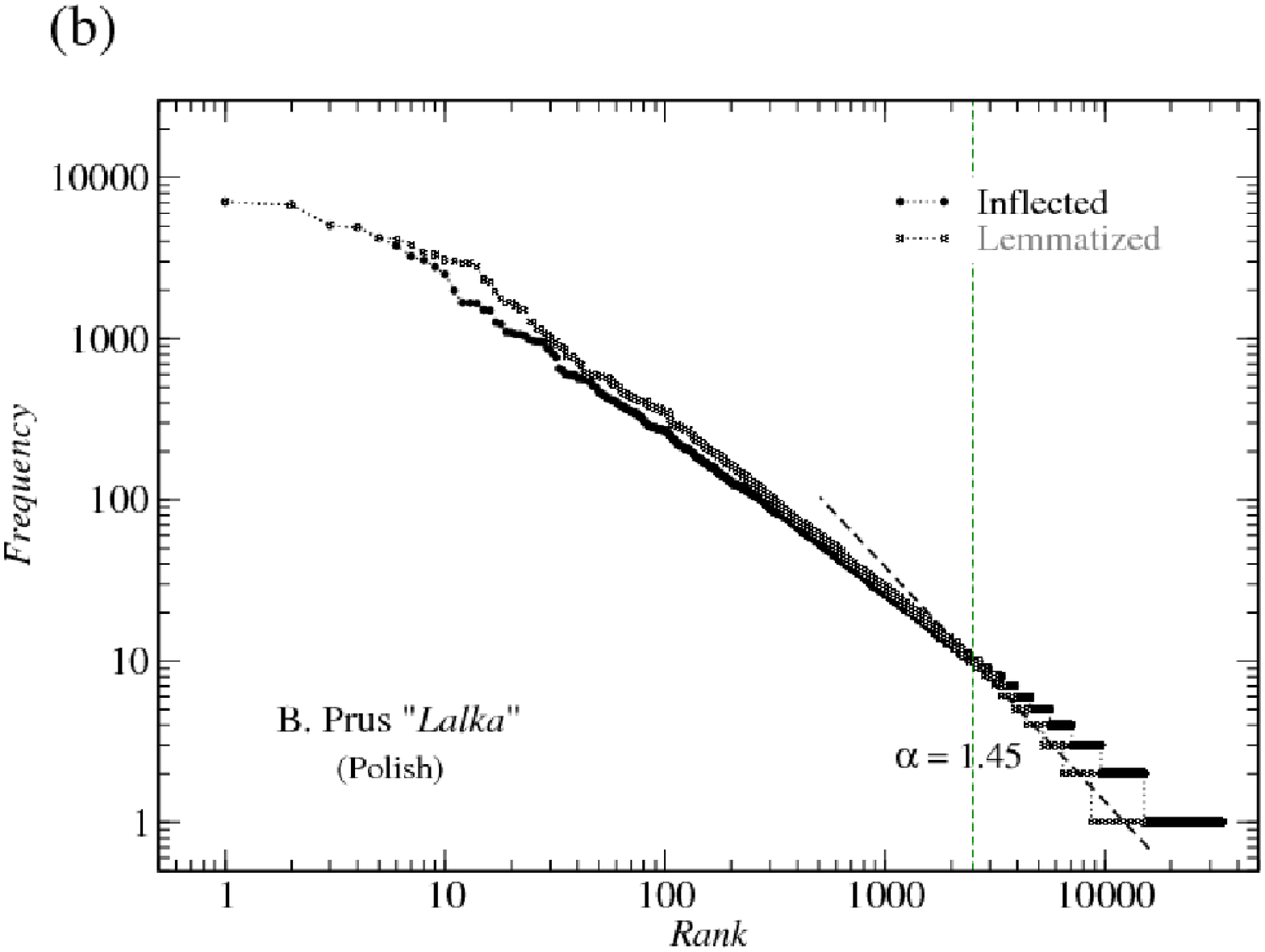}
\caption{Rank-frequency distributions of words in original inflected forms (black circles) together with their lemmatized versions (grey squares) for English ``Ulysses'' (a) and Polish ``Lalka'' (b). Crossover points between the preserved scale-invariant regime for lower ranks and its faster-decaying part for higher ranks corresponding to lemmatized words are denoted by vertical dashed lines. Power-law functions fitted for the high-rank regimes are also shown (dashed lines).}
\end{figure}

Until this point, we have considered only the original (inflected) forms of words as they occured in texts. Therefore the same dictionary item from an inflectable class typically appeared as a few different words in the rankings. Now we also consider lemmas. In Figure 3 we compare the rank-frequency distributions for the inflected and the lemmatized forms of words for ``Ulysses'' (Figure 3(a)) and ``Lalka'' (Figure 3(b)). For both texts, the lemmas show distribution which for higher ranks decays faster than the one for the inflected forms, but the Polish text is more downward-deflected than the English one. This is not surprising, since collapsing the inflected forms into lemmas impoverishes the vocabulary (understood as a set of character sequences) and removes words which predominantly occupy higher ranks. This procedure affects Polish words more than English ones since the former are much more inflectable than the latter. What is interesting, however, is that for both languages there are regions, extending over approximately two orders of magnitude (up to 1,000 in ``Ulysses'' and up to 2,500 in ``Lalka''), in which scale invariance is clear. Inspired by this, a tempting direction for future research could be examining an idea that perhaps not only the word frequency is subject to scale invariance, but also the concepts (i.e. objects, actions, and properties), as they are considered by the brain, do. However, this idea requires a lot more studies to be better grounded.

\begin{figure}
\hspace{-0.5cm}
\epsfxsize 7cm
\epsffile{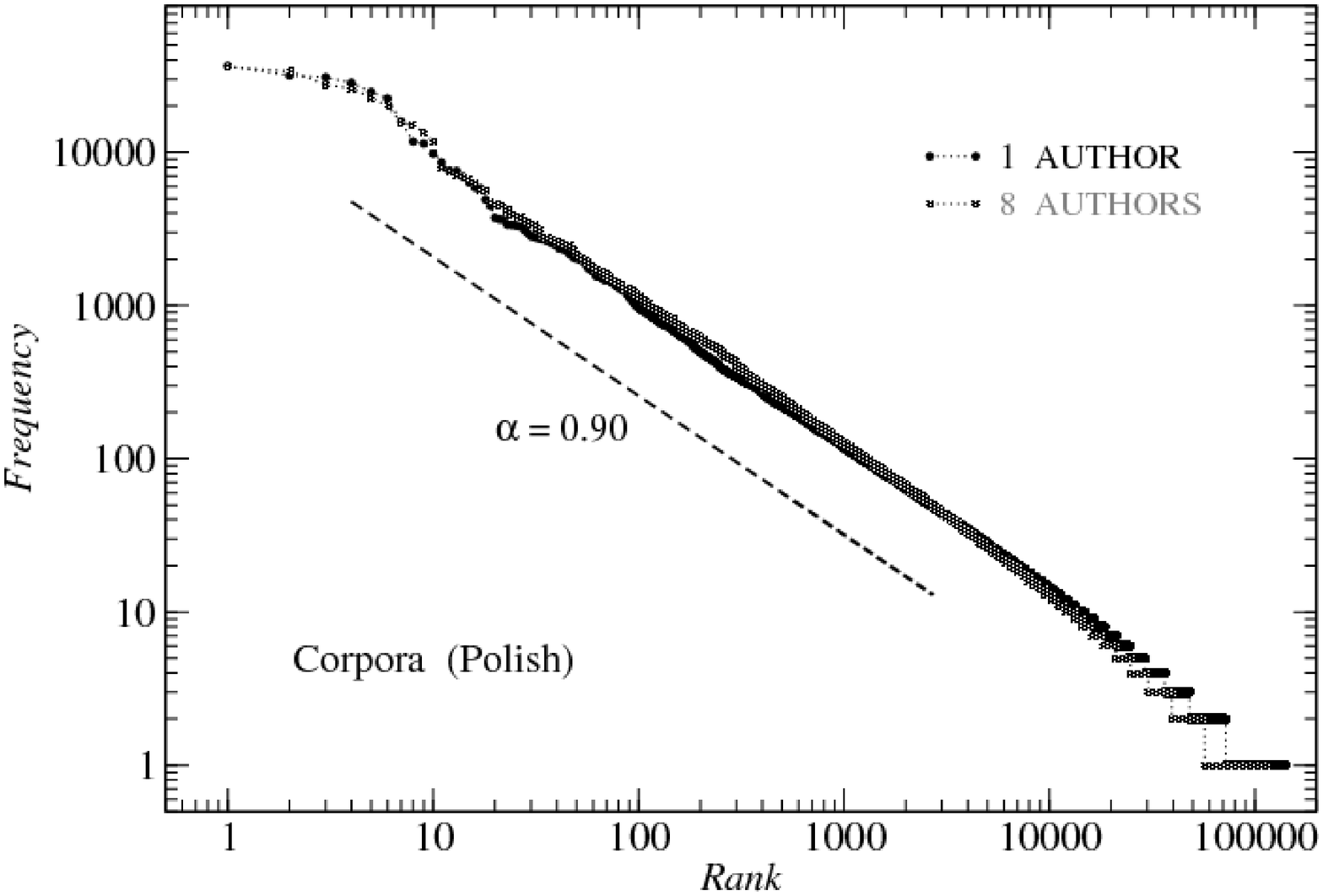}
\epsfxsize 7cm
\epsffile{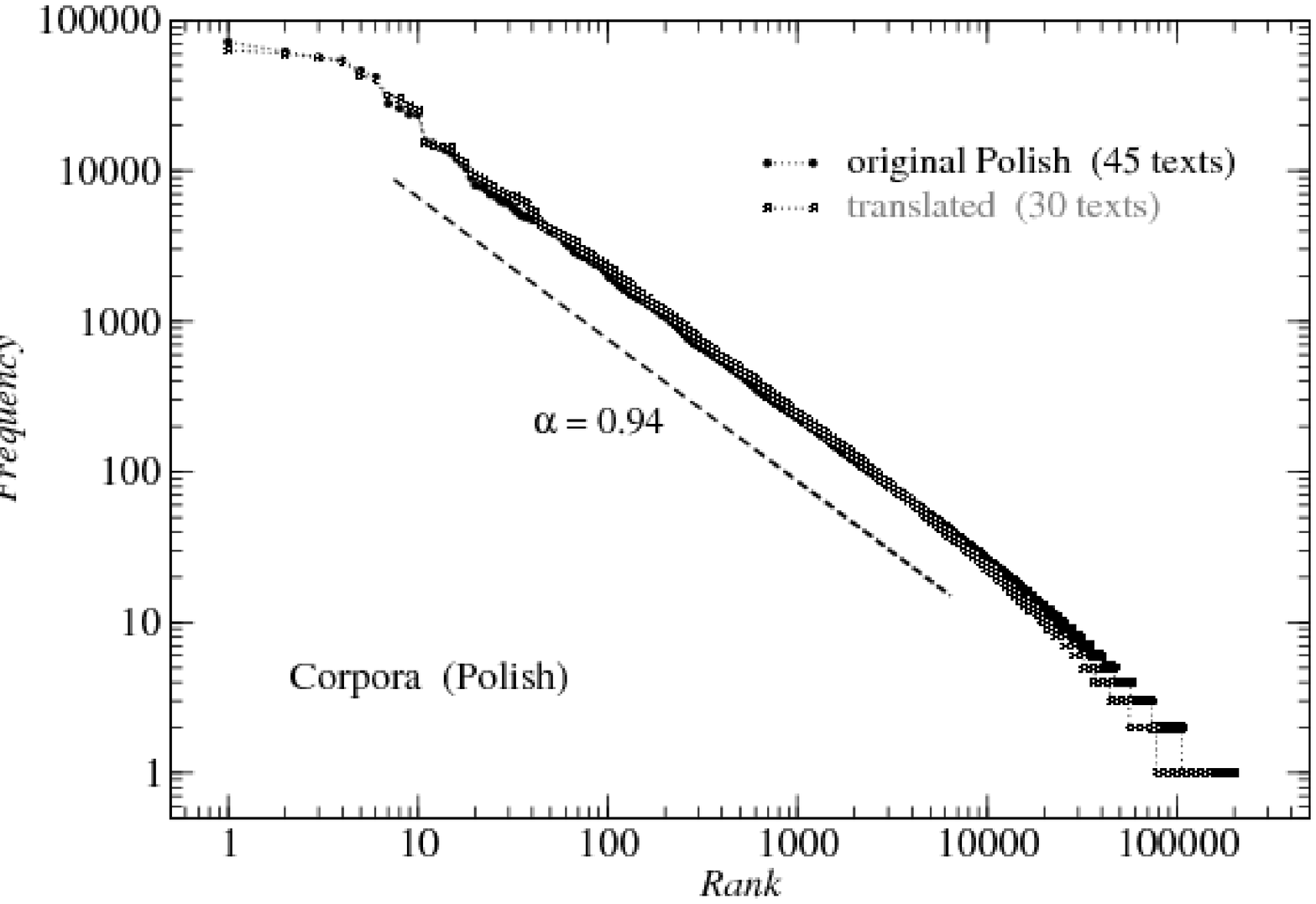}
\caption{Rank-frequency distributions of words in Polish corpora. (a) Corpus consisting of works of the same author (black circles) vs. corpus with works of different authors (grey squares). (b) Corpus consisting of 45 original texts written in Polish (black circles) vs. corpus consisting of 30 foreign texts translated into Polish (grey squares). Power laws with best-fitted exponents: $\alpha \simeq 0.90$ (a) and $\alpha \simeq 0.94$ within the ranks 10-2000 (b) are denoted by dashed lines.}

\end{figure}

As was already mentioned above, passing from a single English text to a larger literary corpora is associated with reaching the limits of the applicability of the Zipf law in its classical form with the exponent $\alpha \simeq 1$. Typically, after a short cross-over range, for ranks larger than a few thousands another scale-invariant regime is observed with $\alpha > 1$~\cite{montemurro01,ferrer01}. Presence of two distinct exponents in the rank-frequency distribution of English words can be explained by the existence of two sets of words: the first one comprises common words which are frequently used by all the authors (thus forming the language core), and the second one comprises the remaining words among which are technical words, words typical for a specific author or words which are otherwise rarely used. However, we propose another complimentary explanation of the breaking of the Zipf law for higher word ranks. Based on books of a few different authors we observe that the Zipf law is better realized for single texts than it is for corpora, even if we consider a corpus to be a collection of works of the same author. Figure 4(a) shows the rank-frequency distributions of words for two small corpora of Polish texts: the first one (black symbols) comprises 26 novels and stories by Polish fantasy writer Andrzej Sapkowski, while the second one (green symbols) is formed of 41 novels and stories written by 8 different authors. The texts in the second corpus were selected in such a way that the total length of each corpus is comparable (1.3 million words). It can easily be noted that for ranks larger that 6,000-8,000 the distribution for the second corpus shows a slightly faster decay than the distribution for the first corpus. In turn, the distribution for the first corpus seems to deviate from the unique scale-invariant behaviour more than any single member text of this corpus (not shown here, but compare this with the result for a single novel in Figure 1(c)). This conclusion, however, must be treated with care since single texts have much smaller vocabulary than larger corpora.

Finally, let us look once again at Figure 1(a) and Figure 1(b), where the Zipf plots for an English text and its Polish translation are presented. Both distributions show an undistorted power-law slope for the whole range of ranks, which means that the scale-invariant character of ``Ulysses'' was preserved by the translator. Actually, if one takes into account the peculiar character of this novel, especially the unusually rich vocabulary, this result has to be considered remarkable. Motivated by this observation, we more systematically look at the distribution of words from texts which were translated into Polish from other languages. We find that although such texts show scale-invariant behaviour for the smallest ranks, for larger ones a breakdown of scale invariance occurs and we see a deflection towards smaller frequencies. In order to compare properties of the translated and the native Polish texts, we constructed the following corpora: the first one consisting of 45 texts written originally in Polish, and the second one consisting of 30 translated texts. Both corpora have similar size of 2.3 million words. The results are presented in Figure 4(b). For the ranks $< 2000$ both corpora develop roughly the same distributions with $\alpha \simeq 0.94$. However, for higher ranks the distribution for the translated corpus decreases faster than the distribution for the native corpus. The observed difference is sufficiently significant to consider it as an actual property of both groups of texts.

\section{Conclusions}

We studied the rank-frequency properties of words in texts written in two languages with different grammar: English (represented by ``Ulysses'') and Polish (represented by the novel ``Lalka''). We considered both the inflected and the lemmatized word forms and in both texts we tagged the words with their parts of speech. We found that the lemmatized words form rankings which decay faster than their counterparts for the inflected words. This difference seems to be higher for Polish words, which on average have more inflected forms than the English ones. Despite this differences, in both languages the rankings exhibit scale invariance. However, this is not true for individual word classes, for which the rank-frequency plots typically are not scale-invariant, with a trace of such behaviour being identifiable only for verbs. Based on the observation that the frequency of words belonging to the same class is not Zipf-like distributed, it is extremely interesting to reiterate that the global rank-frequency distribution of all the words belonging to all the classes is Zipf-like. This result can reflect a complex organization of human language, in which parts do not simply add up to the global picture. The famous statement of P.W.~Anderson~\cite{anderson}, ,,More is different", seems adequate in this case, indeed. From this angle, the linguistic complexity can also manifest itself through the logic of mutual ,,dressing" among words belonging to different parts of speech that the entire proportions emerge scale-invariant even though in majority of these parts separately the proportions do not respect such a kind of organization.

We showed that the rank-frequency distribution of words in a corpus of native Polish texts decays slightly slowler than its counterpart for foreign texts translated into Polish. In the same manner, we compared the rank-frequency plots for individual texts of the same author and for a corpus comprising all these texts. We observed that the plot for the corpus deviates more from the uniform power-law than the ones for individual texts do. In turn, the corpus consisting of texts by different authors decays faster than the corpus for a single author. These evidences suggests that the long-range autocorrelations originating from a given book's continuous narration or a given author's style can considerably influence the usage of words and have impact on their statistical properties. These correlations are distorted if we form a corpus consisting of different works, in the same manner as the unique correlations which are allowed to exist in each particular realization of a system are suppressed if one forms a statistical ensemble from a number of different realizations of this system. This possibility is somehow opposed to the traditional model of analysis in quantitative linguistics, according to which the corpora, due to their larger size, are more useful subject of analysis than single works. In our opinion this leads to losing a significant amount of information on the language structure, which is based, among others, on the correlations. From this point of view, the corpora mixing different texts and averaging out possible correlations specific to single texts, can say less about the complexity of language than individual works.


\begin{thebibliography}{99}

\bibitem{nowak00} M.A.~Nowak, J.B.~Plotkin, V.A.A.~Jansen, Nature {\bf 404}, 495-498 (2000)

\bibitem{zipf49} G.K.~Zipf, "Human behavior and the principle of least effort", Addison-Wesley (Cambridge, 1949)

\bibitem{mandelbrot54} B.~Mandelbrot, Word {\bf 10}, 27 (1954)

\bibitem{ferrer03} R.~Ferrer Cancho, R.V.~Sol\'e, Proc.~Natl.~Acad.~Sci.~USA {\bf 100}, 788-791 (2003)

\bibitem{miller57} G.A.~Miller, Amer.~J.~Psychol. {\bf 70}, 311-314 (1957)

\bibitem{montemurro01} M.A.~Montemurro, Physica A {\bf 300}, 567-578 (2001)

\bibitem{slomczynski} J.~Joyce, "Ulisses", translated by M.~S{\l}omczy\'nski, Wydawnictwo Pomorze (Bydgoszcz, 1992)

\bibitem{bnc} The British National Corpus website: http://www.natcorp.ox.ac.uk/

\bibitem{leech} G.~Leech, P.~Rayson, A.~Wilson, "Word Frequencies in Written and Spoken English: based on the British National Corpus", Longman (London, 2001)

\bibitem{ferrer01} R.~Ferrer Cancho, R.V.~Sol\'e, J.~Quant.~Linguistics {\bf 8}, 165-173 (2001)

\bibitem{anderson} P.W.~Anderson, Science {\bf 177}, 393-396 (1972)

\end{thebibliography}
\end{document}